\documentclass[10pt, a4paper]{article}

\usepackage[final]{lrec2026} 

\usepackage{graphicx}
\usepackage{amsmath}
\usepackage{amsfonts}
\usepackage{algorithm}
\usepackage{algpseudocode}
\usepackage{multirow}
\usepackage{xcolor}
\usepackage{enumitem}
\usepackage{float}

\usepackage{times}
\usepackage{latexsym}

\definecolor{electricpurple}{rgb}{0.75, 0.0, 1.0}
\definecolor{LightSeaGreen}{rgb}{0.2, 0.5, 0.9}
\definecolor{pinkalh}{rgb}{0.858, 0.188, 0.478}

\title{WISTERIA: Weak Implicit Signal-based Temporal Relation Extraction with Attention}

\name{Duy Dao Do, Anaïs Halftermeyer, Thi-Bich-Hanh Dao} 

\address{
University of Orléans, INSA Centre Val de Loire, LIFO EA 4022, France\\
         \{duy-dao.do, anais.halftermeyer, thi-bich-hanh.dao\}@univ-orleans.fr\\}

\abstract{
Temporal Relation Extraction (TRE) requires identifying how two events or temporal expressions are related in time. Existing attention-based models often highlight globally salient tokens but overlook the pair-specific cues that actually determine the temporal relation.
We propose WISTERIA (Weak Implicit Signal-based Temporal Relation Extraction with Attention), a framework that examines whether the top-$K$ attention components conditioned on each event pair truly encode interpretable evidence for temporal classification. Unlike prior works assuming explicit markers such as before, after, or when, WISTERIA considers signals as any lexical, syntactic, or morphological element implicitly expressing temporal order.
By combining multi-head attention with pair-conditioned top-$K$ pooling, the model isolates the most informative contextual tokens for each pair.
We conduct extensive experiments on TimeBank-Dense, MATRES, TDDMan, and TDDAuto, including linguistic analyses of top-$K$ tokens. Results show that WISTERIA achieves competitive accuracy and reveals pair-level rationales aligned with temporal linguistic cues, offering a localized and interpretable view of temporal reasoning.
 \\ \newline  }

\begin{document}

\maketitleabstract

\section{Introduction}

Temporal Relation Extraction (TRE) is a key task in natural language processing (NLP) that identifies and classifies temporal relationships between events and temporal expressions in text. These relationships reveal temporal dynamics such as event order, duration, and causality. Understanding them aids information retrieval, event prediction, and knowledge graph construction. A main challenge in TRE is identifying which contextual elements encode the temporal relationship between two entities, as this directly affects accuracy and interpretability.

Traditional frameworks such as TimeML \citep{pustejovsky2006timebank} introduce \textit{signal words} (e.g., \textit{before}, \textit{after}, \textit{when}) to mark explicit temporal links. Yet, such markers appear sparsely in real-world text. Annotators frequently infer relations using implicit linguistic cues, often through syntactic or common-sense reasoning. For example, in the TimeBank-Dense corpus \citep{cassidy2014annotation}, the pairs \textit{(takeover, news)} and \textit{(spent, thought)} are connected by explicit signals (\textit{before}, \textit{when}), while \textit{(spent, sold)} or \textit{(said, sold)} require implicit reasoning from context or syntax. This observation motivates our hypothesis that many temporal relations are carried by \textit{implicit linguistic signals} beyond explicit markers.

\begin{figure}[ht]
    \centering
    \includegraphics[width=\columnwidth]{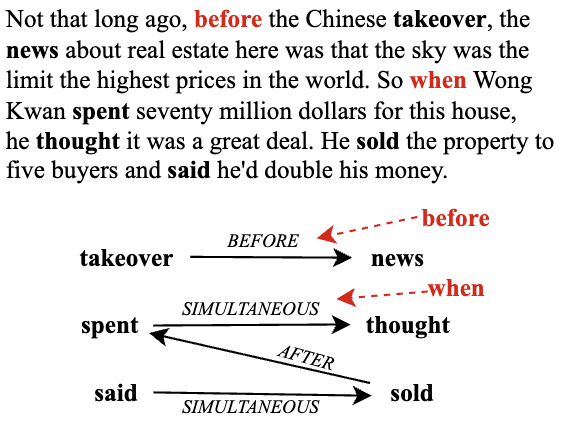}
    \caption{An example from TimeBank-Dense \citep{cassidy2014annotation}, based on TimeBank 1.2 \citep{pustejovsky2006timebank}. Explicit signals link \textit{BEFORE(takeover, news)} and \textit{SIMULTANEOUS(spent, thought)}, while \textit{spent-sold} depends on implicit contextual cues.}
    \label{fig:example}
\end{figure}

Recent works in TRE leverage pretrained language models (PLMs) \citep{lin-etal-2019-bert, Man_Ngo_Van_Nguyen_2022}, graph reasoning \citep{zhang-etal-2022-extracting, mathur-etal-2021-timers, zhao-etal-2021-effective}, logic-based inference \citep{zhou2020clinicaltemporalrelationextraction, huang2023classificationunifiedframeworkevent}, and external knowledge integration \citep{ning-etal-2019-improved, han-etal-2020-domain}. Although effective, these methods are computationally costly and provide limited transparency into how contextual cues drive predictions. Few studies explicitly examine whether attention mechanisms can reveal the contextual rationale behind a predicted relation for each entity pair.

In this work, we introduce \textbf{WISTERIA}, a lightweight and interpretable framework that shifts attention from global salience modeling to \textbf{relation-conditioned evidence modeling}. Instead of computing attention uniformly across the sentence, WISTERIA explicitly conditions attention on each entity pair and retrains only the top-$K$ contextual tokens most influential.

Importantly, our goal is not to claim that attention constitutes a complete explanation of model decisions. Rather, we investigate whether relation-conditioned attention distributions align with systematically meaningful temporal features. To this end, we introduce a structured linguistic interpretability framework that analyzes the part-of-speech (POS), dependency (Dep), and morphological (Morph) properties of the selected top-$K$ tokens. This analysis allows us to examine whether the extracted evidence corresponds to known temporal phenomena such as tense marking, aspectual morphology, and subordinating constructions.

We evaluate WISTERIA on four benchmark datasets: \textbf{TimeBank-Dense} \citep{cassidy2014annotation}, \textbf{MATRES} \citep{ning-etal-2019-improved}, and the discourse-level corpora \textbf{TDDMan} and \textbf{TDDAuto} \citep{Naik2019TDDiscourseAD}. Experimental results demonstrate competitive performance, particularly on sentence-level datasets, while providing consistent linguistic alignment between attention-selected evidence and temporal cues.

Our contributions are threefold:
\begin{itemize}
    \item We introduce a \textbf{relation-conditioned top-$K$ attention mechanism} that models attention as entity-pair-specific evidence selection rather than global token salience.
    \item We propose a \textbf{linguistic interpretability protocol} that systematically analyzes POS, dependency, and morphological distributions of attention-selected tokens to evaluate temporal signal alignment.
    \item We demonstrate that \textbf{WISTERIA} achieves competitive performance across four TRE benchmarks while maintaining computational efficiency and offering structured, linguistically grounded interpretability.
\end{itemize}

\section{Related Work}
\label{sec:Related_work}
Temporal Relation Extraction has evolved through multiple paradigms, including pretrained language models (PLMs), graph-based reasoning, logic-based inference, and knowledge-enhanced approaches \citep{lin-etal-2019-bert, zhang-etal-2022-extracting, mathur-etal-2021-timers, ning-etal-2019-improved, huang2023classificationunifiedframeworkevent}. While these methods achieve strong performance, they often require complex architectures or structured inference pipelines, limiting interpretability and computational efficiency.

\paragraph{Attention in TRE.}
Attention mechanisms \citep{bahdanau2014neural, vaswani2017attention} are widely used in Transformer-based TRE models. However, conventional attention typically distributes weights according to sentence-level salience, which may dilute pair-specific signals. Selective context modeling and graph attention \citep{Man_Ngo_Van_Nguyen_2022, zhang-etal-2022-extracting} improve performance but do not explicitly analyze whether attention aligns with linguistic temporal cues.

\paragraph{Top-$K$ Attention and Evidence Selection.}
Sparse and top-$K$ attention mechanisms have been proposed for efficiency and interpretability in NLP \citep{correia2019adaptively, child2019generating}. In relation extraction, top-$K$ evidence selection has been applied at the document level \citep{ma-etal-2023-dreeam, YUAN2025101728}. However, these approaches operate globally and are not conditioned on specific entity pairs.

\paragraph{Our Positioning.}
WISTERIA differs by explicitly conditioning attention on entity-pair semantics and analyzing the linguistic properties of the selected evidence. By combining relation-conditioned top-$K$ attention with structured linguistic analysis, our framework bridges selective evidence modeling and interpretable temporal reasoning.

\section{Preliminaries}
\label{sec:Preliminaries}

\subsection{Background}

Attention mechanisms \citep{bahdanau2014neural, vaswani2017attention} have become a cornerstone of modern NLP by enabling models to dynamically focus on contextually relevant parts of the input. This mechanism lies at the heart of the Transformer architecture \citep{vaswani2017attention}, which forms the foundation for most state-of-the-art pretrained language models. Transformer-based models such as BERT and RoBERTa \citep{devlin-etal-2019-bert, liu2019robertarobustlyoptimizedbert} leverage attention to generate rich contextual embeddings that capture deep semantic and syntactic dependencies between tokens. However, these models remain constrained by fixed input length (typically 512 tokens) and limited interpretability-attention weights do not always correspond to meaningful linguistic evidence \citep{Jain2019AttentionIN, wiegreffe2019attentionexplanation}.  

To improve efficiency and interpretability, several variants such as multi-head and top-$K$ attention \citep{correia2019adaptively, child2019generating} have been proposed. While top-$K$ attention sparsifies focus by selecting the most relevant tokens, it generally operates globally without conditioning on specific relational pairs. Motivated by this limitation, we introduce a more selective and interpretable mechanism, pair-conditioned top-$K$ attention, built upon the BERT base architecture to enhance contextual focus at the entity-pair level.

\subsection{Attention Pooling}
To obtain a single contextual embedding from token representations $X \in \mathbb{R}^{n \times d}$, we use a simple attention-based pooling mechanism-motivated by prior uses of attention for sequence summarization \citep{9423033, Safari2020SelfattentionEA} - defined as
\begin{equation}
    \label{eq:attpool}
    att\_pooling(X) = Att(m_X, X, X),
\end{equation}
where $m_X$ is the mean of $X$ used as the query. This instantiation highlights tokens most relevant to the context with complexity $O(nd)$. We then extend this idea to pair-conditioned top-$K$ pooling, which selectively aggregates tokens most informative for each entity pair to improve interpretability and efficiency.

\section{Methodology}
\label{sec:Methodology}

\begin{figure*}[ht]
  \includegraphics[width=\linewidth]{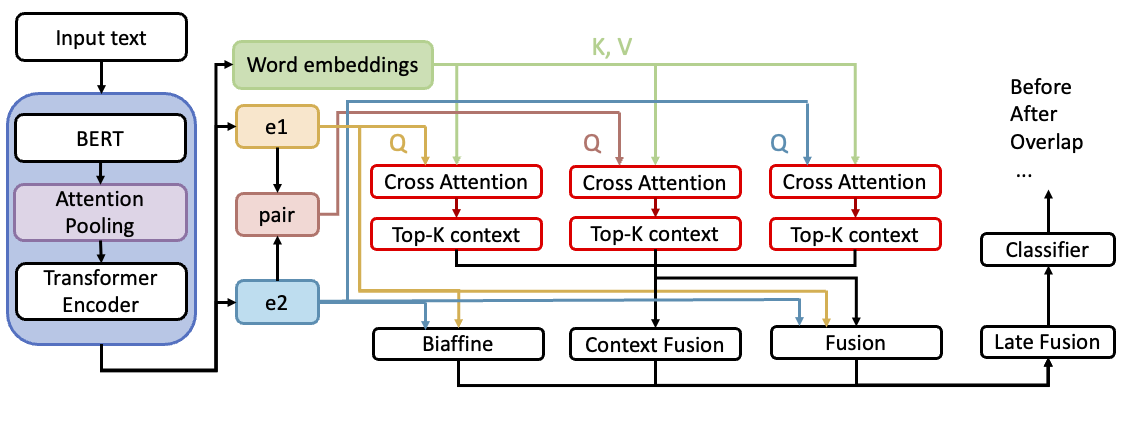}
  \caption{Architecture of WISTERIA. BERT and a transformer encoder generate contextualized representations, while pair-conditioned top-$K$ cross-attention extracts key context for each entity pair. The biaffine, context, and pair-fusion outputs are integrated via late fusion for temporal relation classification.}
  \label{fig:flow}
\end{figure*}

Our proposed framework combines contextual representation learning with selective attention to improve temporal relation extraction. Figure~\ref{fig:flow} illustrates the overall architecture, consisting of four main components: context construction, multi-level embedding extraction, pair-conditioned top-$K$ attention, and relation classification.

\subsection{Context Construction}
Given two entities $e_1$ and $e_2$ in a document, we dynamically extract a context window $C$ using a predefined window size $ws$. When the entities are close, a single centered window is used; when they are far apart, two subwindows centered on each entity are concatenated, ensuring both entities appear within $C$. Each entity span is marked with four special tokens: $[E1], [/E1], [E2], [/E2]$, following prior temporal relation works \citep{ning-etal-2019-improved, lin-etal-2019-bert, dligach-etal-2017-neural, 10.1093/jamia/ocae059}. 

The marked text is tokenized by both a Transformer-based tokenizer (BERT or RoBERTa) and the \textit{spaCy} tokenizer \citep{honnibal-johnson-2015-improved}. Subword-to-token alignment is performed using the spaCy \texttt{Alignment} module, enabling the construction of (i) \textbf{entity masks} for $e_1$ and $e_2$, and (ii) \textbf{word-level masks} linking spaCy tokens to BERT subwords. These masks guide the subsequent attention-based pooling operations.

\subsection{Multi-level Embedding Extraction}
Let $C = [w_1, w_2, \ldots, w_n]$ denote the token sequence. We first obtain subword embeddings $X \in \mathbb{R}^{n \times d}$ from a pretrained BERT encoder.  
To derive word-level embeddings, we apply an attention-based pooling mechanism \citep{9423033, Safari2020SelfattentionEA} over the subwords belonging to each spaCy token:
\begin{equation}
    h^w_i = Att(m_i, X, X),
\end{equation}
where $m_i$ is the mean vector of subwords of word $i$ used as the query.  
Similarly, entity-level embeddings $h_{e_1}$ and $h_{e_2}$ are obtained by aggregating the word embeddings within each marked entity span using the same formulation.

The resulting word embeddings are further refined through a lightweight Transformer encoder with few layers of multi-head self-attention and learnable positional encoding \citep{shaw-etal-2018-self}, producing contextualized representations $H^c = [h^{c}_1, \ldots, h^{c}_M]$.

\subsection{Pair-conditioned Top-$K$ Attention}

Given the entity-level representations $h_{e_1}$ and $h_{e_2}$, we first construct a relation-aware pair representation by concatenation followed by linear projection:
\begin{equation}
    h_{pair} = [h_{e_1} \oplus h_{e_2}] W_p,
\end{equation}
where $W_p$ projects the concatenated entity embedding into the contextual space. This projection aligns the pair representation with contextual word representations, enabling relation-specific attention computation.

\paragraph{Relation-conditioned cross-attention.}


Unlike conventional attention mechanisms that distribute weights based on sentence-level salience, our approach conditions attention distributions on entity-pair semantics. Specifically, we instantiate cross-attention using three query representations - $e_1$, $e_2$, and the projected pair embedding $h_{pair}$ - following the multi-head attention formulation of \citet{vaswani2017attention}. For each query $q \in \{e_1, e_2, pair\}$, attention scores are computed over contextual representations $H_c$. Because each query encodes entity- or pair-specific semantics, the resulting attention distributions are relation-dependent rather than sentence-global, thereby functioning as pair-aware evidence selectors rather than generic token importance estimators.



\paragraph{Top-$K$ evidence selection.}
To promote selectivity, we retain only the top-$K$ contextual tokens with the highest attention weights for each query, following sparse attention principles \citep{correia2019adaptively, child2019generating}. Let $H^{k}_q$ denote the selected tokens for query $q$, which are aggregated as:
\begin{equation}
    h^{k}_{q} = Att(h'_q, H^{k}_q, H^{k}_q),
\end{equation}
where $h'_q$ is the projected query representation.

Because selection is conditioned on entity-pair semantics, different pairs within the same sentence may focus on distinct contextual evidence. This contrasts with global top-$K$ attention, where token selection is relation-independent, and enables pair-specific evidence modeling.



\paragraph{Label-aware gating.}
To enhance relational discriminability, we introduce learnable label embeddings \citep{ma2016label, rios2018few} as semantic prototypes for temporal relation classes. Instead of serving solely as output targets, these embeddings are incorporated into the representation space and fused with contextual features through a gating mechanism. This allows the model to modulate pair representations according to class-level semantics. The label embeddings act as soft guidance rather than hard constraints, complementing relation-conditioned evidence selection while preserving architectural simplicity.

\subsection{Relation Classification}
We employ three complementary prediction heads:  
(1) a biaffine scorer \citep{dozat2017deepbiaffineattentionneural} modeling direct interactions between $e_1$ and $e_2$;  
(3) a context head focusing solely on top-$K$ contextual signals; and
(2) a fusion head combining entity and contextual features. 
Their logits are integrated using a learnable late-fusion vector:
\begin{equation}
    \hat{y} = \text{softmax}\left( \sum_{i=1}^{3} w_i \cdot \text{logits}_i \right),
\end{equation}
and the entire model is optimized using the cross-entropy objective.  

Overall, our architecture achieves computational complexity $O(L(n^2 d + d^2))$, comparable to Transformer-based encoders, while yielding more selective and interpretable attention distributions.

\section{Experiments}
\label{sec:Experiments}
The experiments aim to answer the following research questions:
\begin{itemize}[noitemsep, topsep=5pt]
\item \textbf{Q1}: How effective is our model in capturing temporal relations compared to standard BERT-based baselines?
\item \textbf{Q2}: To what extent is our model interpretable in revealing how contextual signals contribute to temporal reasoning?
\end{itemize}

\subsection{Datasets \& Evaluation}
We evaluate our model on four benchmark datasets for temporal relation extraction. \textbf{TimeBank-Dense (TBD)} \citep{cassidy2014annotation} densifies the original TimeBank corpus, adding more temporal links than human annotators would naturally provide. MATRES \citep{ning-etal-2019-improved} simplifies TBD by reducing relation classes and removing weakly grounded links (not on the same "temporal axis"), and adding it to the \textbf{AQUAINT} \citep{Graff2002Aquaint} corpus, while \textbf{TDDMan} and \textbf{TDDAuto} \citep{Naik2019TDDiscourseAD} 
are new versions of TBD, introducing inter-sentential relations for discourse-level reasoning, respectively in a manual way, and automatically.

Table~\ref{tab:datasets} summarizes the data distribution and label sets for all four datasets.

\begin{table}[h!]
\centering
\scriptsize
\setlength{\tabcolsep}{8pt}
\begin{tabular}{lcccc}
\hline
\textbf{Dataset} & \textbf{Train} & \textbf{Validation} & \textbf{Test} & \textbf{Labels} \\
\hline
TDDMan & 4000 & 650 & 1500 & a, b, s, i, ii \\
TDDAuto  & 32609 & 1435 & 4258 & a, b, s, i, ii \\
MATRES*  & 10404 & 1836 & 817 & e, a, b, v \\
TBD  & 4032 & 629 & 1427 & a, b, s, i, ii, v \\
\hline
\end{tabular}
\caption{Train/Validation/Test data distribution for TDDMan, TDDAuto, MATRES, and TimeBank-Dense. 
Label abbreviations: a = \textit{After}, b = \textit{Before}, s = \textit{Simultaneous}, i = \textit{Includes}, ii = \textit{Is\_included}, 
v = \textit{Vague}, e = \textit{Equal}. 
(*\citep{ning-etal-2019-improved} use TimeBank and Aquaint for training, Platinum for testing, and 20\% of the training data as validation.)}
\label{tab:datasets}
\end{table}

For all datasets, we report the standard Precision, Recall, and Micro-average F1 scores to ensure consistent comparison across models.  
Following common practice in temporal relation extraction, for TimeBank-Dense and MATRES we exclude the Vague label from evaluation, as done in previous studies \citep{huang2023classificationunifiedframeworkevent, mathur-etal-2021-timers, zhang-etal-2022-extracting, 9515702, Yao_2024}.

\subsection{Baseline Models}
We compare our proposed model (\textbf{WISTERIA}) with a comprehensive set of strong baselines from recent work on temporal relation extraction across  four benchmark datasets.  
The compared methods include:  
(1) \textbf{BiLSTM} \citep{Cheng2017}, an LSTM-based architecture with discourse-level context;  
(2) fine-tuned \textbf{BERT-based} models as implemented in \citet{Ballesteros2020};  
(3) 
\textbf{Unified Framework} \citep{huang2023classificationunifiedframeworkevent}, combining pretrained language models with logical or structural constraints;  
(4) graph-based methods such as \textbf{TIMERS} \citep{mathur-etal-2021-timers}; 
(5) graph-distillation and contrastive learning approaches including \textbf{MuLCo} \citep{Yao_2024};  
(6) the \textbf{DTRE} model \citep{wang-etal-2022-dct} for discourse-level reasoning; and  
(7) \textbf{CPTRE} \citep{YUAN2024111410}, a contrastive pretraining model for document-level temporal relation extraction.  

Our model differs from these baselines by incorporating a pair-conditioned top-$K$ cross-attention mechanism, which selectively focuses on the most informative contextual signals for each entity pair, enhancing both performance and interpretability.

\subsection{Experimental Settings}
We fine-tune the \textbf{BERT-base-uncased}\footnote{\url{https://huggingface.co/google-bert/bert-base-uncased}} model \citep{devlin-etal-2019-bert} with an additional single Transformer encoder layer to derive contextualized embeddings. The model is implemented in PyTorch using the fused AdamW optimizer \citep{loshchilov2017decoupled}. All experiments are conducted on a single NVIDIA Quadro GV100 GPU. Table~\ref{tab:hyperparams} summarizes the key hyperparameters used in our experiments.

\begin{table}[!ht]
\centering
\small
\setlength{\tabcolsep}{5pt}
\begin{tabular}{lc}
\hline
\textbf{Setting} & \textbf{Value} \\
\hline
Pretrained model & BERT-base-uncased \\
Encoder layer & 1 Transformer layer \\
Optimizer & AdamW (fused) \\
Dropout rate & 0.5 \\
Learning rate & $1 \times 10^{-5}$ \\
Batch size & 128 \\
Epochs & 30 \\
Attention heads & 8 \\
Top-$K$ range & [1-20] \\
FFN hidden size & 512 \\
Activation & GELU \\
Hardware & 1 × Quadro GV100 GPU \\
\hline
\end{tabular}
\caption{Hyperparameter configuration for training WISTERIA.}
\label{tab:hyperparams}
\end{table}

\subsection{Results}
\begin{table*}[!ht]
\centering
\resizebox{\textwidth}{!}{
\begin{tabular}{lccccc}
\hline
\textbf{Model} & \textbf{PLM} & \textbf{TBD} & \textbf{MATRES} & \textbf{TDDAuto} & \textbf{TDDMan} \\
\hline
\textbf{BiLSTM} \citep{Cheng2017} & word2vec & 0.484 & 0.595 & 0.518 & 0.243 \\
\hline
\textbf{BERT-based} \citep{Ballesteros2020} & BERT-base & 0.622 & 0.772 & 0.623 & 0.375 \\
\hline
\textbf{Unified-Framework} \citep{huang2023classificationunifiedframeworkevent} & RoBERTa-Large & 0.681 & 0.826 & -- & -- \\
\hline
\textbf{TIMERS} \citep{mathur-etal-2021-timers} & BERT-Large & 0.678 & 0.823 & 0.711 & 0.455 \\
\textbf{DTRE} \citep{wang-etal-2022-dct} & BERT-base & 0.692 & -- & 0.702 & 0.500 \\
\textbf{MuLCo} \citep{Yao_2024} & BERT-base & 0.814 & \textbf{0.857} & 0.662 & 0.473 \\
\textbf{CPTRE} \citep{YUAN2024111410} & BERT-base & 0.714 & 0.842 & \textbf{0.807} & \textbf{0.568} \\
\hline
\textbf{WISTERIA (Ours)} & BERT-base & \textbf{0.831} & 0.843 & 0.709 & 0.4973 \\
\hline
\end{tabular}
}
\caption{Comparison of F1-scores across four benchmark datasets: TimeBank-Dense (TBD), MATRES, TDDAuto, and TDDMan. 
The best performance for each dataset is highlighted in bold.}
\label{tab:results}
\end{table*}

Table~\ref{tab:results} presents the F1-scores across four benchmark datasets. WISTERIA archives strong performance on sentence-level datasets (TBD and MATRES) and competitive results on discourse-level datasets (TDDAuto and TDDMan). Compared with early neural architectures such as BiLSTM, the model demonstrates substantial gains, confirming the benefit of contextualized representations combined with relation-conditioned evidence selection.

Against standard BERT-based baselines, WISTERIA yields notable improvements (e.g., +20.9 F1 on TBD) despite relying on a lightweight architecture with only one additional Transformer layer. When compared with graph-based or constraint-driven systems (TIMERS, DTRE, MuLCo, CPTRE), our model achieves comparable performance while maintaining architectural simplicity. This suggests that relation-conditioned attention provides an efficient mechanism for capturing temporal cues without requiring heavy structural modules.

Across datasets, WISTERIA performs best on TBD and MATRES, which primarily contain short-range, sentence-level relations. On document-level corpora, performance remains competitive but trails graph-based models that explicitly propagate long-distance dependencies. This distinction highlights the modeling scope of WISTERIA: it specializes in localized relational evidence selection rather than global constraint enforcement.

\begin{figure}[!ht]
    \begin{center}
  \includegraphics[width=\columnwidth]{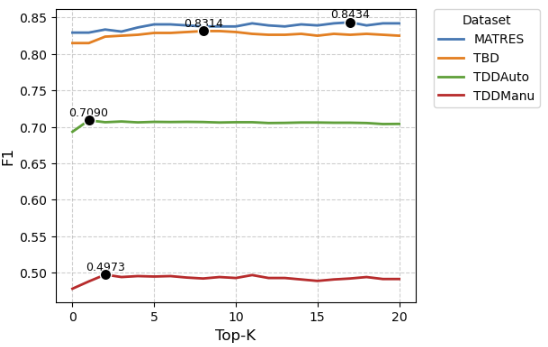}
  \caption{Effect of Top-$K$ values on F1 performance across four datasets.}
  \label{fig:topk}
  \end{center}
\end{figure}
Figure~\ref{fig:topk} illustrates the variation of F1 scores with different Top-$K$ values across datasets. Despite variations in $K$ across corpora, the stable F1 scores suggest notable differences in discourse nature. TBD and MATRES form a coherent group-unsurprising since MATRES includes TBD-as both involve short-distance relations, whereas TDD (Man and Auto) mainly exhibit long-distance ones. Figure \ref{tab:entity_distance_stats} shows the distance between entity pairs across temporal datasets.

\begin{table*}[!ht]
\centering
\resizebox{\textwidth}{!}{
\begin{tabular}{lcccc}
\hline
\textbf{Dataset} & \textbf{Avg. Character Distance} & \textbf{Avg. Token Distance} & \textbf{Min Distance} & \textbf{Max Distance} \\
\hline
TBD      & 116.99 & 19.51 & 1   & 366 \\
MATRES   & 128.94 & 24.01 & 1   & 416 \\
TDDAuto  & 800.99 & 127.61 & 2   & 3349 \\
TDDMan   & 913.61 & 146.65 & 61  & 3074 \\
\hline
\end{tabular}
}
\caption{Average, Minimum, and Maximum Character and Token Distance between Entity Pairs across Temporal Datasets}
\label{tab:entity_distance_stats}
\end{table*}

Overall, WISTERIA remains stable across a wide range of $K$, indicating that the pair-conditioned attention consistently selects informative contextual tokens without overfitting to specific thresholds.  
The optimal performance is achieved $K=8$ for TBD (0.8314) and $K=18$ for MATRES (0.8434), suggesting that moderate context aggregation may be sufficient for capturing most temporal cues in local event pairs.

For document-level datasets (TDDAuto, TDDMan), performance shows minimal improvement beyond $K=[1;2]$, reflecting that adding distant context contributes little additional signal and may even introduce noise due to cross-sentence sparsity.  
This pattern confirms that local temporal relations benefit more from focused attention, while global reasoning requires richer discourse-level modeling beyond token-level selection.

\subsection{Top-$K$ Analysis}
We do not claim that attention weights alone constitute fully faithful or causally complete explanations of model decisions. Rather, we evaluate whether relation-conditioned top-$K$ attention exhibits systematic alignment with linguistically recognized temporal cues. Our analysis therefore focuses on distributional consistency, structural enrichment, and cross-dataset stability, rather than on token-level causal attribution. The distribution of POS, Dependency and Morphologocial features of TimeBank-Dense, MATRES, TDDAuto, TDDMan are shown in Appendix (\ref{appendix}).

\subsubsection{Linguistic Analysis on Local Datasets}
For TimeBank-Dense and MATRES, which primarily contain sentence-level temporal relations, we examine the linguistic composition of the selected top-$K$ tokens.

\paragraph{POS}
In both datasets, nouns (\texttt{NOUN}) and proper nouns (\texttt{PROPN}) dominate the attention space of $e_1$ and $e_2$ about 20\% and 12--13\% respectively-indicating that the model concentrates on event-denoting tokens and their argument structures.  
For the pair-conditioned representation, the attention distribution shifts toward verbs (\texttt{VERB}, $\approx$11--12\%) and adpositions (\texttt{ADP}, $\approx$11--12\%), showing that the model emphasizes functional and relational markers (e.g., "before", "in", "during") when forming a temporal link between two events (TBD, $2572/18936$ tokens); (MATRES, $2151/17016$ tokens). 

This pattern is consistent across TBD and MATRES, confirming that pair-conditioned top-$K$ attention effectively identifies lexical items carrying temporal meaning. 

\paragraph{Dependency}
The dependency distributions reveal distinct syntactic focuses across datasets.
In TBD, temporal entities mainly align with nominal and predicate–argument structures,
dominated by \texttt{amod}, \texttt{compound}, \texttt{det}, \texttt{nsubj}, \texttt{pobj}, and \texttt{prep}
($\approx$7--11\%), reflecting event mentions embedded in noun phrases or governed by prepositions.
In contrast, MATRES shows a similar but more clause-oriented pattern, with
\texttt{compound}, \texttt{det}, \texttt{nsubj}, \texttt{pobj}, \texttt{prep}, and notably \texttt{punct}
as top dependencies ($\approx$8--12\%).
The prominence of \texttt{punct} suggests stronger reliance on clause boundaries and coordination signals.
Across both datasets, the pair-conditioned view amplifies connective dependencies such as
\texttt{mark}, \texttt{advcl}, and \texttt{prep}, which frequently introduce temporal clauses
(e.g., \emph{when}, \emph{as}, \emph{after}) and facilitate pair-level temporal reasoning.

\paragraph{Morphological}
Morphological evidence reveals that \textsc{Wisteria} is sensitive to tense and aspectual cues. Tokens marked with \texttt{Tense=Past} or \texttt{Aspect=Perf} occur frequently across all entity types (around 20--22\% in TBD and 17--18\% in MATRES), indicating that the model leverages verbal morphology to infer temporal ordering, particularly relations such as \textit{BEFORE} and \textit{AFTER}. In the pair-conditioned context, the distribution of \texttt{Tense=Past} and \texttt{Tense=Pres} becomes more balanced, suggesting that the model compares events situated in different temporal frames rather than focusing on a single clause’s internal tense. Overall, these patterns confirm that the top-$k$ tokens are not randomly attended but encode distinct morphosyntactic evidence relevant to temporal reasoning.
\begin{figure*}[!ht]
    \begin{center}
        \includegraphics[width=\linewidth]{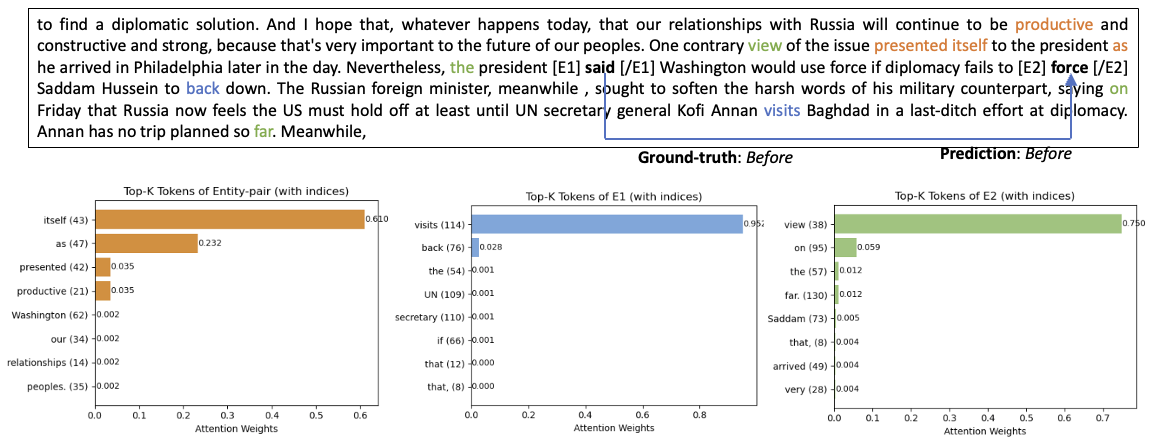}
        \caption{Example from the TimeBank-Dense test set illustrating pair-conditioned top-$K$ attention. The model predicts \textit{BEFORE} between \texttt{said} ($E_1$) and \texttt{force} ($E_2$). Pair-level attention highlights connective cues (e.g., \texttt{as}, \texttt{itself}), while entity-level attention anchors event-specific tokens (e.g., \texttt{visits}, \texttt{Saddam}). The complementary patterns indicate structured evidence selection for temporal inference.}
        \label{fig:top-k_example}
    \end{center}
\end{figure*}
\subsubsection{Linguistic Analysis on Global Datasets}
We extend our linguistic analysis to the two document-level subsets of the TDDiscourse corpus: TDDAuto and TDDMan. These datasets contain long-range and cross-sentence temporal relations, where explicit lexical signals are scarcer and temporal reasoning depends more on discourse-level structures.

\paragraph{POS}
The POS distributions of TDDAuto and TDDMan reveal a stronger reliance on nominal categories and a relative reduction in overt relational markers.  
Across both datasets, nouns (\texttt{NOUN}) are the most frequent category ($\approx$19-21\%), followed by proper nouns (\texttt{PROPN}, $\approx$12-17\%).  
These proportions are notably higher than those of verbs (\texttt{VERB}, $\approx$10-12\%) and adpositions (\texttt{ADP}, $\approx$10-11\%), showing that the model grounds temporal reasoning primarily in event anchors and discourse entities rather than explicit lexical connectors.  
The pair-conditioned attention shows modest increases in \texttt{VERB} and \texttt{ADP}, but overall the token distribution remains noun-dominant-consistent with the more implicit temporal cues found in document-level narratives.

\paragraph{Dependency}
Dependency patterns further confirm this shift toward discourse-level inference.  
For $e_1$ and $e_2$, core dependencies such as \texttt{nsubj}, \texttt{obj}, and \texttt{amod} occur most frequently ($\approx$7--10\%), indicating that attention centers on syntactic heads and event arguments.  
In the pair-conditioned representation, temporal relations are often mediated through \texttt{prep}, \texttt{advcl}, and \texttt{relcl} ($\approx$9--11\%), suggesting that the model selectively captures structural links between clauses or sentences.  
Compared to TBD and MATRES, however, the frequency of overt subordinating markers (e.g., \texttt{mark}, \texttt{advcl}) is lower, implying that WISTERIA must rely on more implicit contextual dependencies to infer relations across discourse boundaries.

\paragraph{Morphological}
At the morphological level, both TDDAuto and TDDMan exhibit diverse tense-aspect patterns reflecting narrative variation.  
Tokens with \texttt{Aspect=Perf|Tense=Past} remain prominent ($\approx$17--19\%), but less dominant than in local datasets, suggesting that the model draws less from explicit verb morphology and more from contextual coherence.  
Forms tagged with \texttt{VerbForm=Inf} ($\approx$4--5\%) and \texttt{VerbForm=Fin} ($\approx$1--2\%) are more frequent, consistent with the narrative reporting style typical of multi-sentence documents.  
Overall, morphological cues contribute less distinctly to temporal ordering here, emphasizing the need for higher-level relational reasoning.

\subsubsection{Summary}
Across all four datasets, the linguistic analyses reveal a clear continuum between local and global temporal reasoning. In sentence-level datasets such as TBD and MATRES, WISTERIA anchors each event ($e_1$, $e_2$) in noun phrases and relies on verbal and prepositional tokens (e.g., \emph{when}, \emph{as}, \emph{after}) together with tense/aspect morphology to establish explicit temporal links. In contrast, in document-level datasets such as TDDAuto and TDDMan, the model encounters fewer overt temporal connectives and instead integrates information from syntactic dependencies (\texttt{compound}, \texttt{amod}, \texttt{prep}) and discourse context that span multiple sentences. This indicates a shift from surface lexical cues to structural and contextual signals of temporal progression. Such transition from explicit to implicit temporal encoding explains why WISTERIA achieves higher F1 scores on local datasets, while maintaining slightly lower yet competitive performance ($\approx$0.70) on global datasets-highlighting both its interpretability and adaptability across different reasoning scopes.

\subsubsection{Interpretability}

Figure~\ref{fig:top-k_example} presents a representative example from TimeBank-Dense. Pair-level attention highlights connective and contextual framing tokens (e.g., \textit{as}, \textit{itself}, \textit{productive}), whereas entity-level attention anchors event-specific roles. The complementary patterns indicate a functional distinction between event grounding and relational signaling.

We do not claim strict causal faithfulness. Rather, our analysis demonstrates structured and linguistically coherent evidence selection. Across datasets, attention distributions show (i) enrichment of temporal morphosyntactic features, (ii) stability across $K$ values, and (iii) consistent specialization between entity-level and pair-level queries.

These findings suggest that relation-conditioned attention functions as a structured evidence filter rather than a diffuse salience mechanism. Establishing formal causal guarantees remains an open challenge, and integrating constraint-based reasoning over top-$K$ evidence is a promising direction for future work.

\section{Conclusion}
\label{sec:Conclusion}

We presented \textbf{WISTERIA}, a lightweight framework for temporal relation extraction that introduces relation-conditioned top-$K$ attention for entity-pair-specific evidence selection. By conditioning attention on event-pair semantics, WISTERIA isolates informative contextual cues and achieves competitive performance across four benchmarks (TimeBank-Dense, MATRES, TDDAuto, and TDDMan) while maintaining computational efficiency.

Beyond predictive accuracy, our linguistic analyses show that the selected top-$K$ tokens consistently align with temporal morphosyntactic features, providing structured and transparent evidence selection. Rather than serving as a global salience mechanism, relation-conditioned attention functions as a pair-aware evidence filter.

Future work will integrate constraint-driven or rule-based reasoning on top of the extracted evidence to extend WISTERIA from pairwise classification toward globally consistent temporal reasoning.

To support reproducibility, we release the full implementation and top-$K$ attention files\footnote{\url{https://github.com/doduydao/WISTERIA}}.

\section{Limitations}
\label{sec:Limitations}

While WISTERIA advances relation-conditioned evidence selection and structured interpretability in temporal relation extraction, several limitations remain.

\paragraph{Independent pairwise modeling.}
WISTERIA models each entity pair independently and does not incorporate explicit temporal constraints such as transitivity, antisymmetry, or global consistency enforcement. As a result, although pair-conditioned top-$K$ attention effectively isolates locally informative contextual signals, it does not guarantee document-level temporal coherence. This limitation is particularly evident in discourse-level datasets (TDDAuto and TDDMan), where long-distance and multi-hop dependencies require structured reasoning beyond token-level evidence selection. Integrating constraint-driven or graph-based inference modules on top of the extracted evidence constitutes an important direction for future work.

\paragraph{Local evidence vs. global reasoning.}
The current top-$K$ mechanism operates at the level of token selection conditioned on a specific entity pair. While this design enhances focus and interpretability, it does not explicitly propagate information across multiple pairs within a document. Consequently, WISTERIA excels at localized relational reasoning but is not designed as a full temporal reasoning engine. Future extensions may explore hybrid architectures that combine relation-conditioned attention with hierarchical, graph-based, or energy-based global reasoning frameworks.

\paragraph{Interpretability scope.}
The interpretability provided by WISTERIA is structural and distributional rather than strictly causal. Our analyses demonstrate systematic alignment between attention-selected tokens and linguistically recognized temporal cues; however, attention weights alone do not constitute provably faithful explanations in the causal sense. Establishing formal guarantees of explanation faithfulness remains a broader open challenge in neural interpretability research. We therefore position WISTERIA’s interpretability as a transparent evidence selection mechanism rather than a complete model of human-understandable reasoning.

\paragraph{Dependence on automatic linguistic annotation.}
The POS, dependency, and morphological analyses rely on automatic annotations, which may introduce noise. Although the consistency of observed patterns across datasets mitigates this concern, future work could investigate tighter integration between attention-selected evidence and rule-based linguistic validation.

Overall, WISTERIA should be viewed as a lightweight, relation-conditioned evidence extraction framework. Extending it with explicit temporal constraints or symbolic reasoning components represents a promising pathway toward unified predictive and reasoning-based temporal understanding.

\section{Ethics statement}
This work did not raise ethical concerns during its development and is not expected to pose any in the future. From an ethical perspective, the contribution of this research lies in advancing toward more resource-efficient and interpretable systems. It is important to note that the knowledge extracted from these systems should not be conflated with human cognitive understanding of linguistic temporality. Rather, insights derived from neural models should be treated as investigative leads to be empirically tested against theories of human temporal processing. Furthermore, from a broader societal perspective, research that advances the interpretability of deep learning systems contributes to enhanced human oversight and control of these increasingly ubiquitous technologies in daily life.

\section{References}\label{sec:reference}

\bibliographystyle{lrec2026_natbib}
\bibliography{lrec2026}

\bibliographystylelanguageresource{lrec2026_natbib}
\bibliographylanguageresource{languageresource}

\section{Appendix}
\label{appendix}

\subsection{POS distribution}

\begin{figure*}[!ht]
    \begin{center}
  \includegraphics[width=\linewidth]{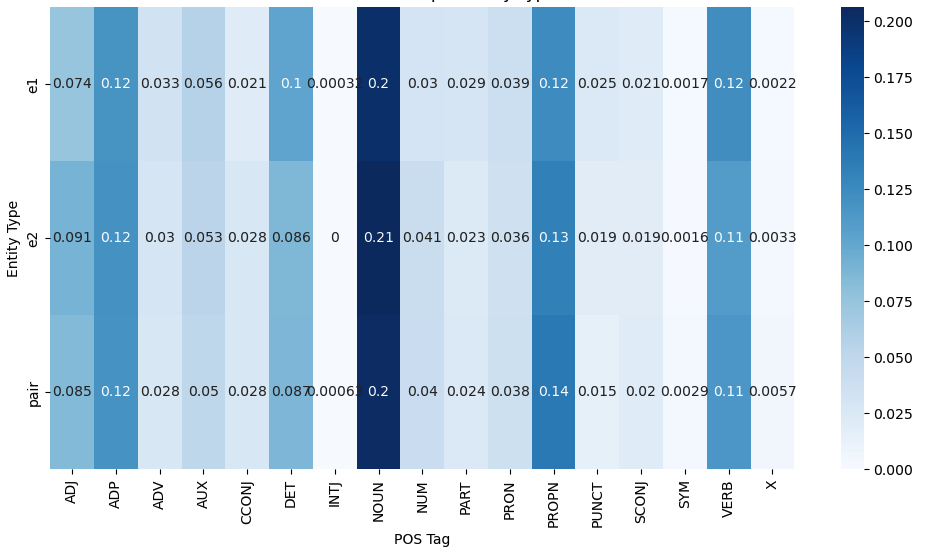}
  \caption{POS feature distribution of TimeBank-Dense}
  \label{fig:POS-TBD}
  \end{center}
\end{figure*}

\begin{figure*}[!ht]
    \begin{center}
  \includegraphics[width=\linewidth]{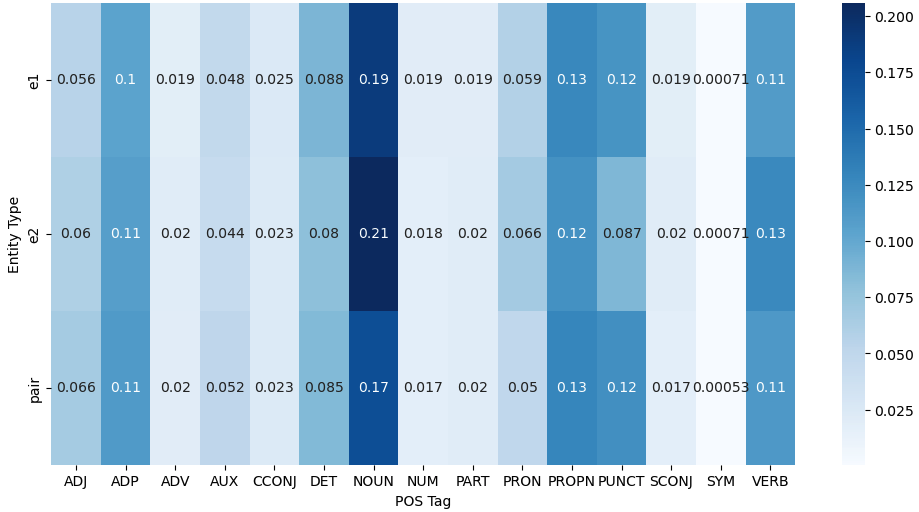}
  \caption{POS feature distribution of MATRES}
  \label{fig:POS-MATRES}
  \end{center}
\end{figure*}

\begin{figure*}[!ht]
    \begin{center}
  \includegraphics[width=\linewidth]{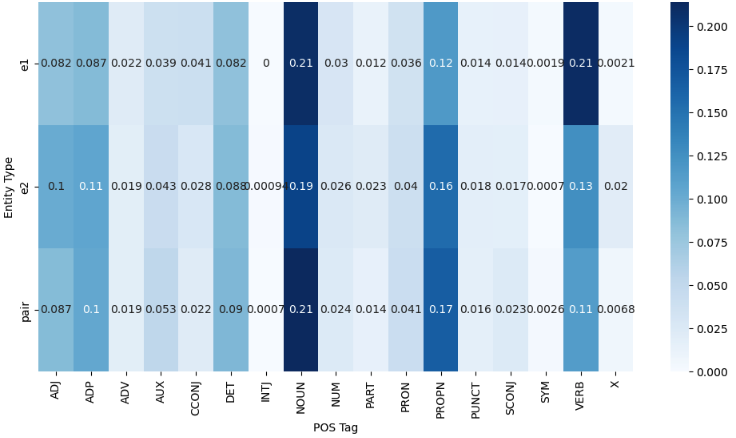}
  \caption{POS feature distribution of TDDAuto}
  \label{fig:POS-TDDAuto}
  \end{center}
\end{figure*}

\begin{figure*}[!ht]
    \begin{center}
  \includegraphics[width=\linewidth]{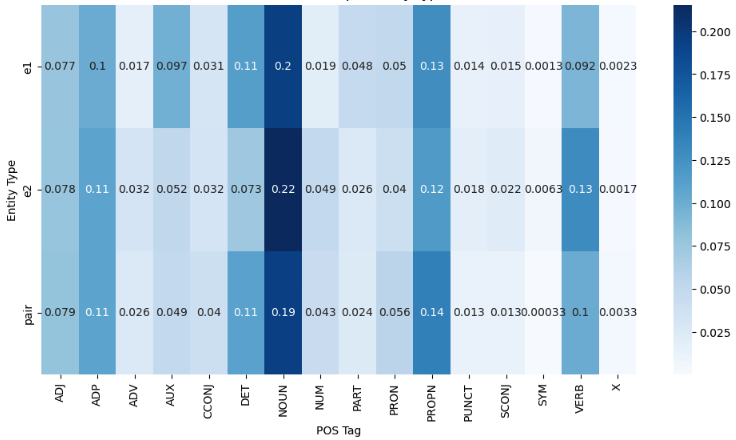}
  \caption{POS feature distribution of TDDman}
  \label{fig:POS-TDDMan}
  \end{center}
\end{figure*}

\subsection{Dependency feature distribution}

\begin{figure*}[!ht]
    \begin{center}
  \includegraphics[width=\linewidth]{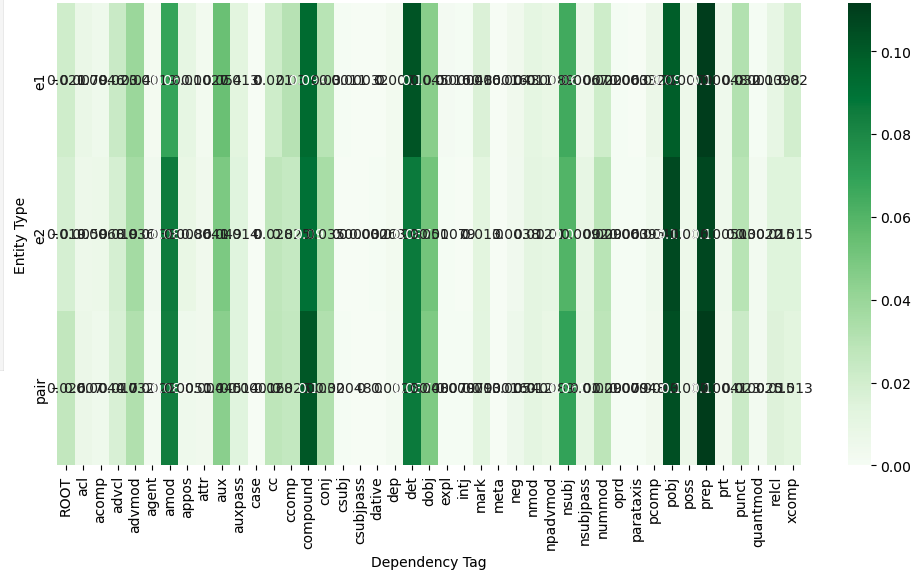}
  \caption{Dependency feature distribution of TimeBank-Dense}
  \label{fig:Dep-TBD}
  \end{center}
\end{figure*}

\begin{figure*}[!ht]
    \begin{center}
  \includegraphics[width=\linewidth]{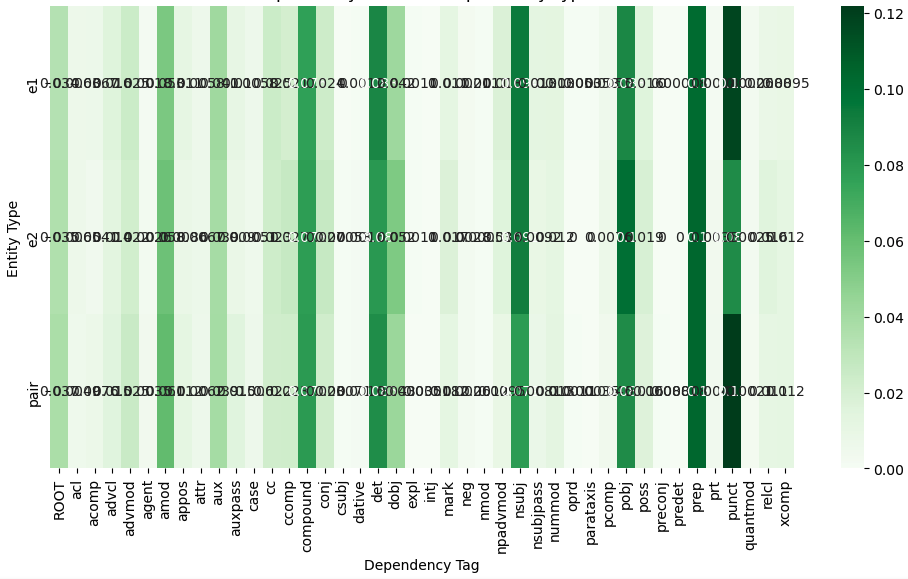}
  \caption{Dependency feature distribution of MATRES}
  \label{fig:Dep-MATRES}
  \end{center}
\end{figure*}

\begin{figure*}[!ht]
    \begin{center}
  \includegraphics[width=\linewidth]{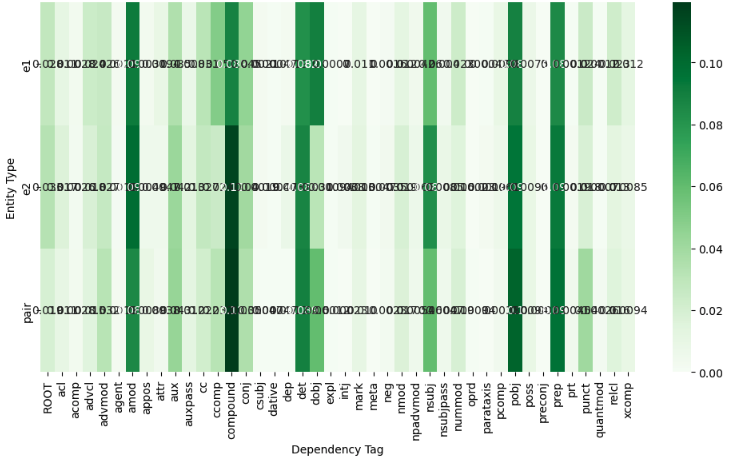}
  \caption{Dependency feature distribution of TDDAuto}
  \label{fig:Dep-TDDAuto}
  \end{center}
\end{figure*}

\begin{figure*}[!ht]
    \begin{center}
  \includegraphics[width=\linewidth]{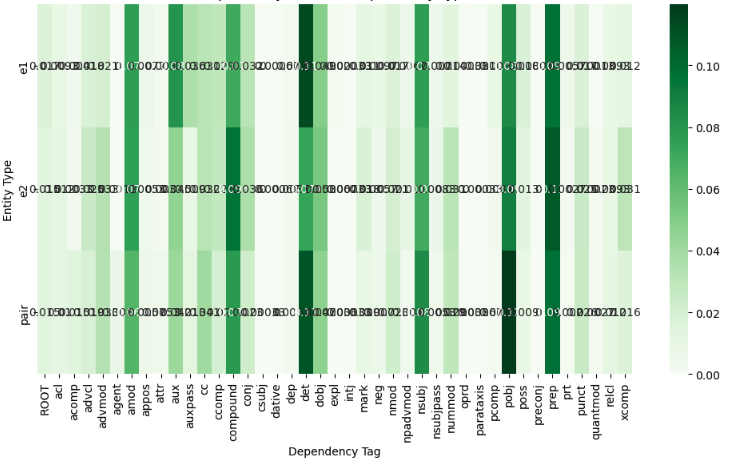}
  \caption{Dependency feature distribution of TDDMan}
  \label{fig:Dep-TDDMan}
  \end{center}
\end{figure*}

\subsection{Morphological feature distribution}
\begin{figure*}[!ht]
    \begin{center}
  \includegraphics[width=\linewidth]{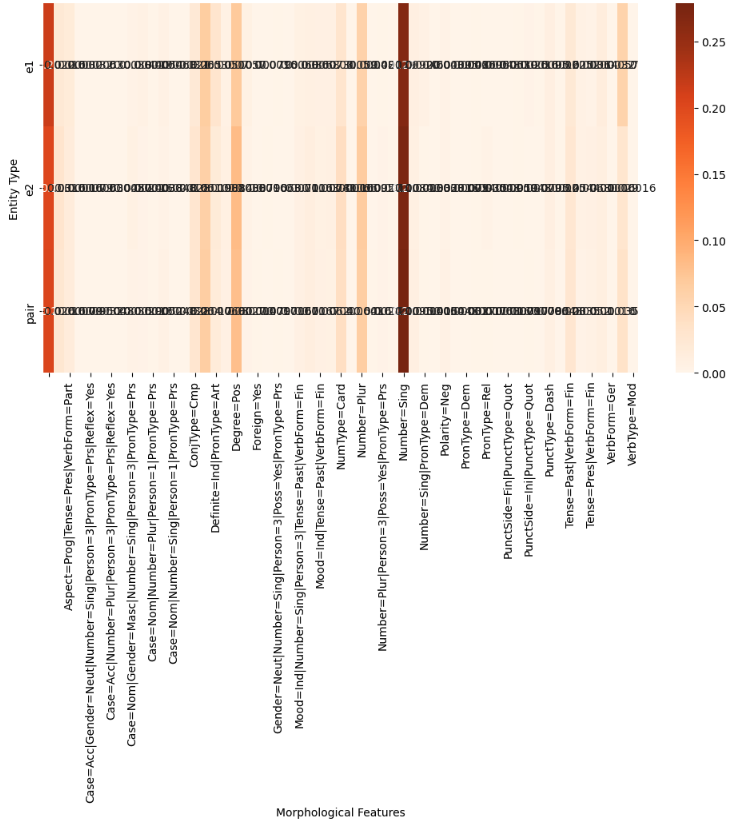}
  \caption{Morphological feature distribution of TimeBank-Dense}
  \label{fig:Morph-TBD}
  \end{center}
\end{figure*}

\begin{figure*}[!ht]
    \begin{center}
  \includegraphics[width=\linewidth]{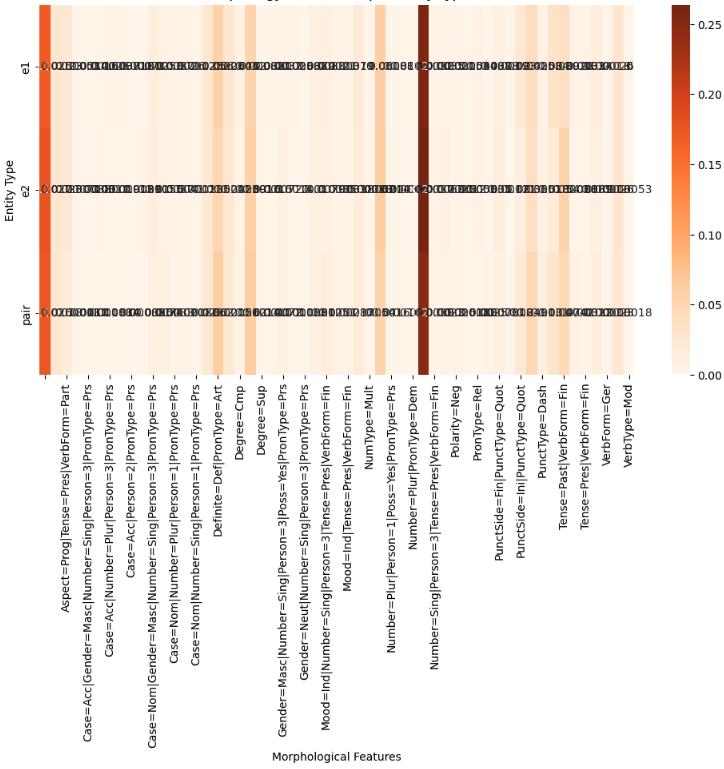}
  \caption{Morphological feature distribution of MATRES}
  \label{fig:Morph-MATRES}
  \end{center}
\end{figure*}

\begin{figure*}[!]
    \begin{center}
  \includegraphics[width=\linewidth]{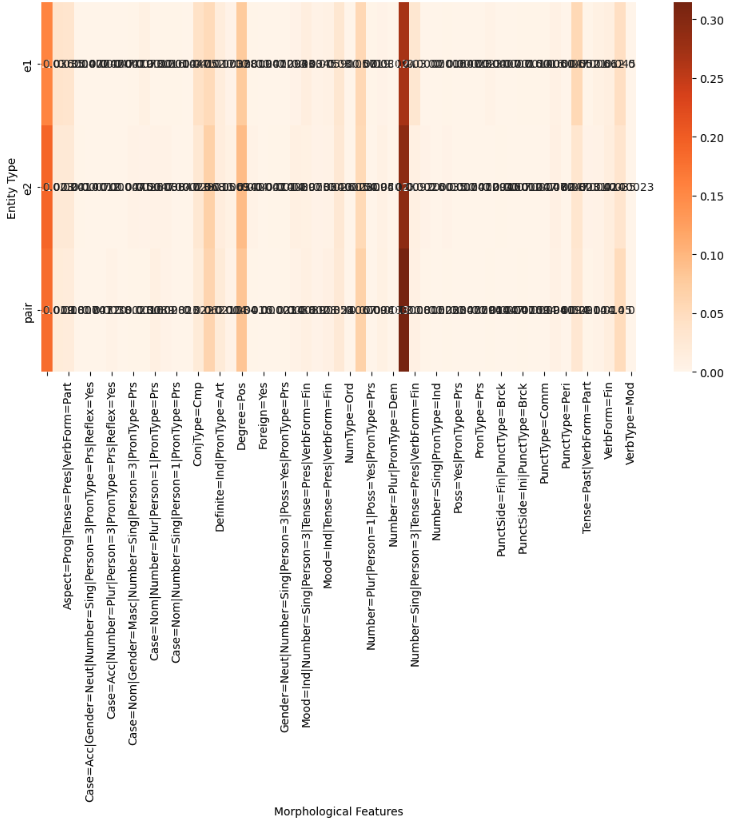}
  \caption{Morphological feature distribution of TDDAuto}
  \label{fig:Morph-TDDAuto}
  \end{center}
\end{figure*}

\begin{figure*}[!ht]
    \begin{center}
  \includegraphics[width=\linewidth]{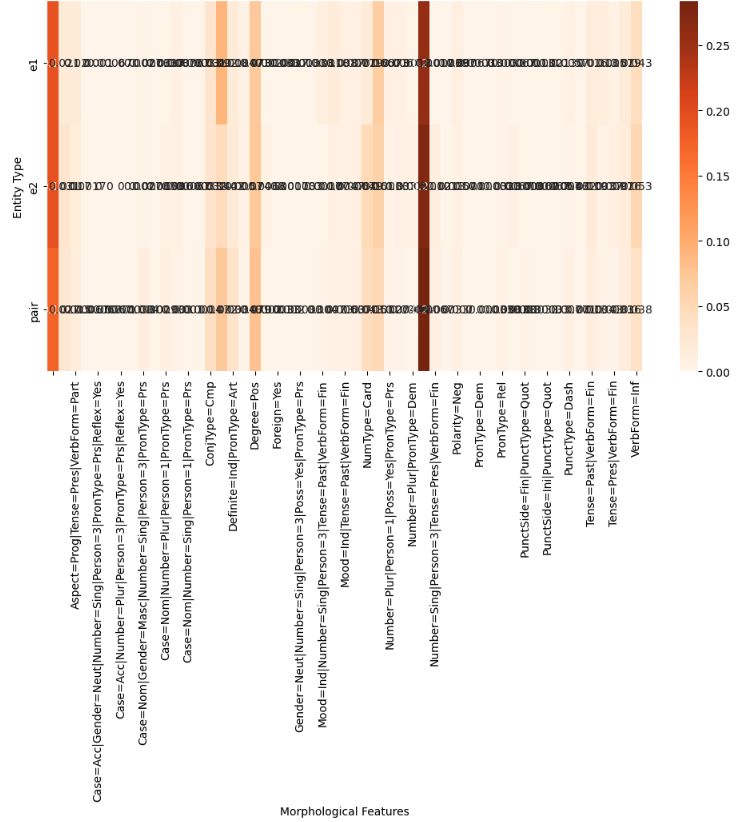}
  \caption{Morphological feature distribution of TDDMan}
  \label{fig:Morph-TDDMan}
  \end{center}
\end{figure*}

\end{document}